\documentclass[letterpaper, 10 pt, conference]{ieeeconf}  
\usepackage{ifxetex}
\ifxetex\else\usepackage[utf8]{inputenc}\fi

\IEEEoverridecommandlockouts                              

\usepackage{subcaption}
\usepackage{amssymb}
\usepackage{amsmath}
\usepackage{cases}
\usepackage{amsfonts}
\usepackage{bm}

\usepackage{textcomp}

\usepackage[capitalise]{cleveref} 
\crefname{section}{Sec.}{Secs.}
\Crefname{section}{Section}{Sections}
\crefname{figure}{Fig.}{Figs.}
\Crefname{figure}{Figure}{Figures}
\crefname{equation}{}{}
\Crefname{equation}{Equation}{Equations}

\usepackage[hyphens,spaces,obeyspaces]{url}
\usepackage{tikzscale}

\usepackage{rotating}

\usepackage[detect-all]{siunitx}
\usepackage{pgfplots}

\usepackage{tikz}
\usepgfplotslibrary{groupplots}
\usetikzlibrary{plotmarks}
\usetikzlibrary{positioning}
\usetikzlibrary{shapes,arrows, fit, positioning}
\usepgfplotslibrary{external}
\usepackage{tikz-3dplot}
\usepackage{tikzscale}
\usepackage{centeredtikzarrowbulletheads}
\tikzset{>=latex}
\tikzset{
 font={\fontsize{10pt}{12}\selectfont}}
\pgfplotsset{every tick label/.append style={font=\small}}
\pgfplotsset{legend style={font=\small}}
\pgfplotsset{every axis label/.append style={font=\small}}

\title{\LARGE \bf
Segmentation of Robot Movements using Position and Contact Forces}

\author{
\centering Martin Karlsson* \quad Anders Robertsson \quad Rolf Johansson
\thanks{* The authors work at the Department of Automatic Control, Lund University, PO Box 118,
SE-221 00 Lund, Sweden.\protect\\
{Martin.Karlsson@control.lth.se\protect\\
The research leading to these results has received funding from the Vinnova project \textit{Kirurgens Perspektiv}, Boverket \textit{Innovativt bostadsbyggande} \#6438/2018, and Vinnova \textit{Uppkopplad byggplats} \#2017-05202. The authors are members of the ELLIIT Excellence Center at Lund University.}}
}

\begin{document}
\newcommand{\cmt}[1]{{\color{red}{\textbf{Comment:} #1}}}

\newtheorem{proposition}{Proposition}
\newtheorem{theorem}{Theorem}
\newtheorem{lemma}{Lemma}
\newtheorem{remark}{Remark}
\newtheorem{definition}{Definition}


\maketitle
\thispagestyle{empty}
\pagestyle{empty}
\begin{abstract}
In this paper, a method for autonomous segmentation of demonstrated robot movements is proposed. Position data is clustered into Gaussian mixture models (GMMs), and an initial set of segments is identified from the Gaussian basis functions. A Kalman filter is used to detect sudden changes in the contact force/torque measurements, and this is used to update and verify the initial segmentation points. The segmentation method is verified experimentally on an industrial robot.
\end{abstract}

\section{Introduction}
Robot programs typically consist of sequences of pre-defined movements, operating in well-structured industrial environments. This concept has proven successful for automating a wide range of manufacturing tasks in, for instance, the automotive industry. However, traditional robot programming is difficult and therefore accessible to experts only. Further, robot programs have low adaptability to changes in the environment. Therefore, robots are rare outside labs and factories, and even seemingly repetitive tasks, such as assembly tasks, are commonly done manually.

To facilitate robot programming, a lot of research has been devoted to develop and evaluate robot programming by demonstration (PbD) \cite{billard2008robot,niekum2015learning,karlsson2019human}. One approach is based on lead-through programming, where a human grasps the robot and moves it along the desired trajectory \cite{pan2010recent,stolt2015sensorless}. Another approach is reinforcement learning (RL) \cite{stulp2012model,levine2013guided,levine2015learning,li2018reinforcement}, where the main obstruction in robotics, as compared to simulated environments such as computer games, is the limited amount of attempts and failures that are feasible on hardware. It would also be possible to combine PbD and RL, by using demonstrated movements and interaction forces as prior information to warm start the RL.

\begin{figure}
\centering
	\begin{tikzpicture}
		\def\px{0.5}
		\def\py{4.5}
		\def\sc{0.65}
	    \node[anchor=south west,inner sep=0] at (0,0) {\includegraphics[width=0.98\columnwidth]{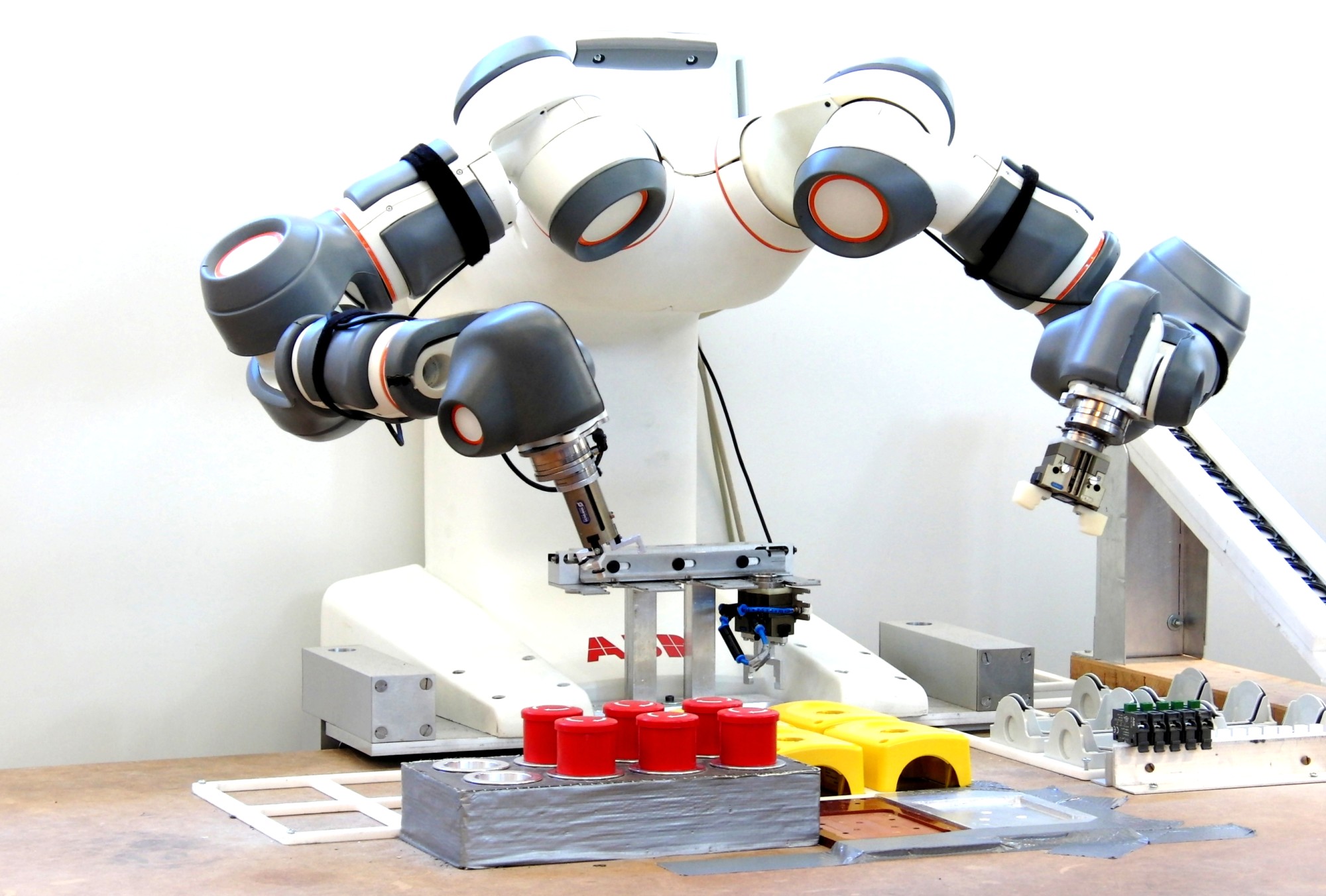}};   
	    \draw[thick,->,>=stealth,red] (\px, \py) -- (\px+1*\sc, \py);
	    \draw[thick,->,>=stealth,black] (\px, \py) -- (\px, \py+1*\sc);
	    \draw[blue,thick] (\px, \py) circle [radius=0.2*\sc];
	    \draw[fill=black] (\px, \py) circle [radius=0.065*\sc];
	    \node[anchor=south,red] at (\px+0.75, \py) {$y$};
	    \node[anchor=west,black] at (\px, \py+0.75)  {$z$};
	    \node[anchor=south,blue] at (\px+0.25, \py-0.4) {$x$};
	\end{tikzpicture}
\caption{The ABB YuMi \cite{yumi} prototype robot used in the experiments. Directions $({\color{blue} x},{\color{red} y},z)$ are indicated in the upper left.}
\label{fig:yumi_main}
\end{figure}
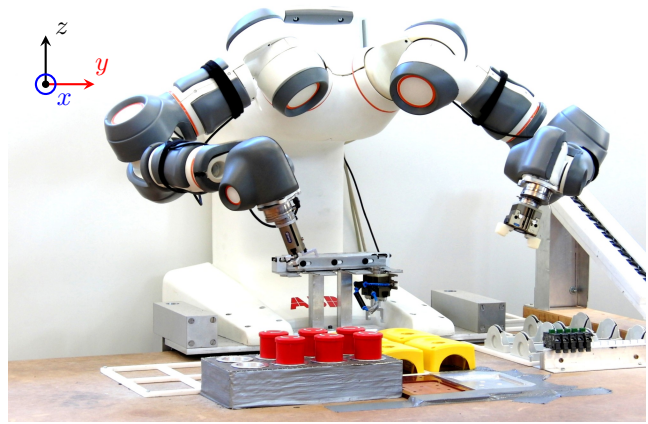

Both in the context of PbD and RL, it is useful to divide an intended task into key phases. The process of dividing a demonstration into such key phases is commonly referred to as segmentation. It has been applied to a wide range of tasks, such as assembly tasks \cite{lioutikov2015probabilistic}, surgery \cite{lin2006towards,krishnan2018transition}, and household tasks \cite{lee2015autonomous}. In PbD, the segmentation is useful for extracting sub-tasks that are semantically meaningful, and allowing for debugging and re-arrangement of the sub-tasks during the programming process. In RL, segmentation could be used to initialize rewards not only at the final goal state, but also at the sub-goals, \textit{i.e.}, upon completion of each key phase, to guide the learning of the task.

In \cite{lee2015autonomous}, demonstrations were represented as position time series, and subsequently clustered into Gaussian Mixture Models (GMMs). Each basis function of a GMM corresponded to one segment, and one segmentation point was assigned between each consecutive basis functions in the time dimension. The main benefit with this approach was that the segmentation was done autonomously, and without requiring manual tuning of parameters. Table~1 in \cite{lee2015autonomous} shows a qualitative comparison of \cite{lee2015autonomous} and related segmentation methods. The method was evaluated using unstructured demonstrations of everyday tasks, such as rice cooking and table setting. The result was semantically meaningful segments, which could be re-arranged to adjust and re-use previous robot programs.

Successful segmentation is important in this context. Failure of obtaining meaningful segments with correct pre- and post-conditions conditions would endanger the subsequent learning, debugging, and execution. Significant position uncertainties in the work space are common in robotic manipulation, and contact forces can then be used to determine whether contact has been established between the robot and a work object. It would therefore be desirable to add force measurements to the method in \cite{lee2015autonomous}.

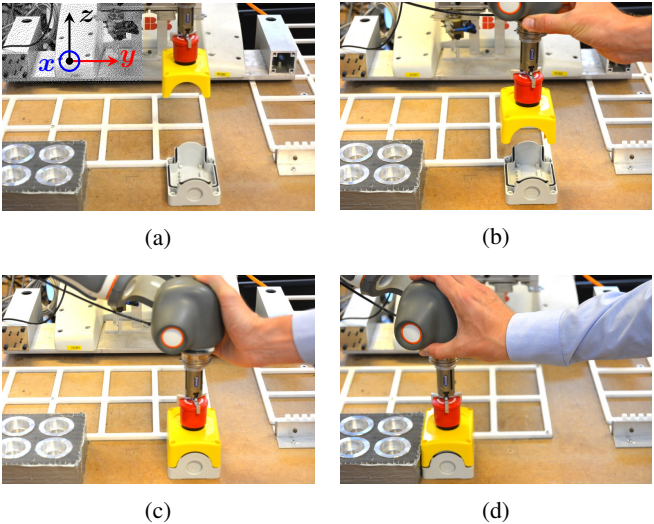
\begin{figure}
\begin{minipage}{.48\columnwidth}
\centering
\begin{tikzpicture}
		\def\px{0.9}
		\def\py{2}
		\def\sc{0.65}
	    \node[anchor=south west,inner sep=0] at (0,0) {\includegraphics[width=\columnwidth]{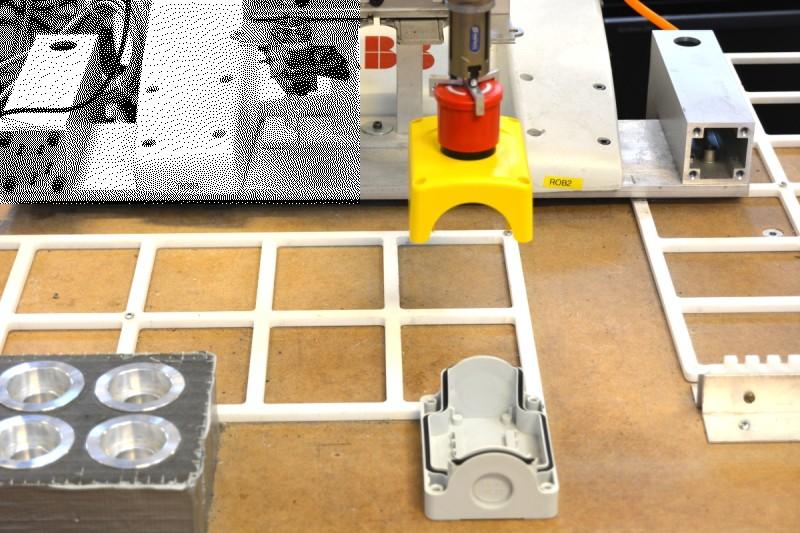}};   
	    \draw[thick,->,>=stealth,red] (\px, \py) -- (\px+1*\sc, \py);
	    \draw[thick,->,>=stealth,black] (\px, \py) -- (\px, \py+1*\sc);
	    \draw[blue,thick] (\px, \py) circle [radius=0.2*\sc];
	    \draw[fill=black] (\px, \py) circle [radius=0.065*\sc];
	    \node[anchor=south,red] at (\px+0.78, \py-.2) {{\boldmath $y$}};
	    \node[anchor=west,black] at (\px, \py+0.6)  {{\boldmath $z$}};
	    \node[anchor=south,blue] at (\px-.3, \py-0.25) {{\boldmath $x$}};
	\end{tikzpicture}
\subcaption{}
\vspace{3mm}
\end{minipage}
\hfill
\begin{minipage}{.48\columnwidth}
\centering
\includegraphics[width=\columnwidth]{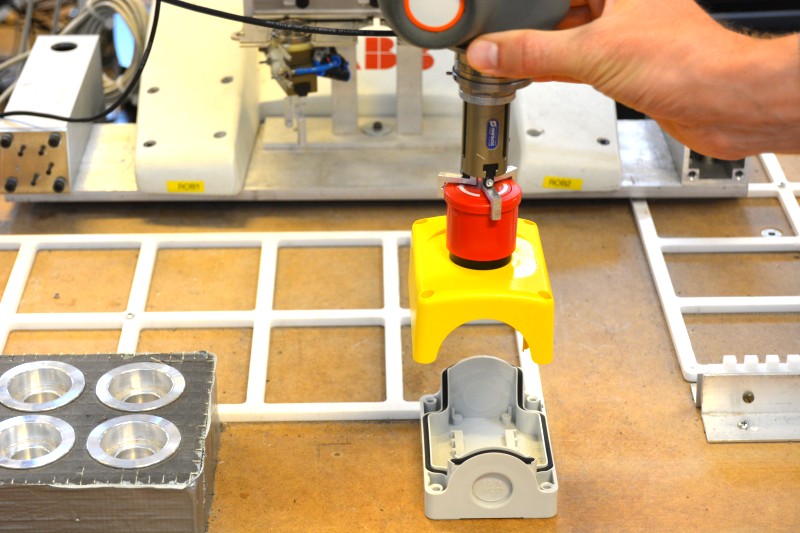}\subcaption{}
\vspace{3mm}
\end{minipage}
\begin{minipage}{.48\columnwidth}
\centering
\includegraphics[width=\columnwidth]{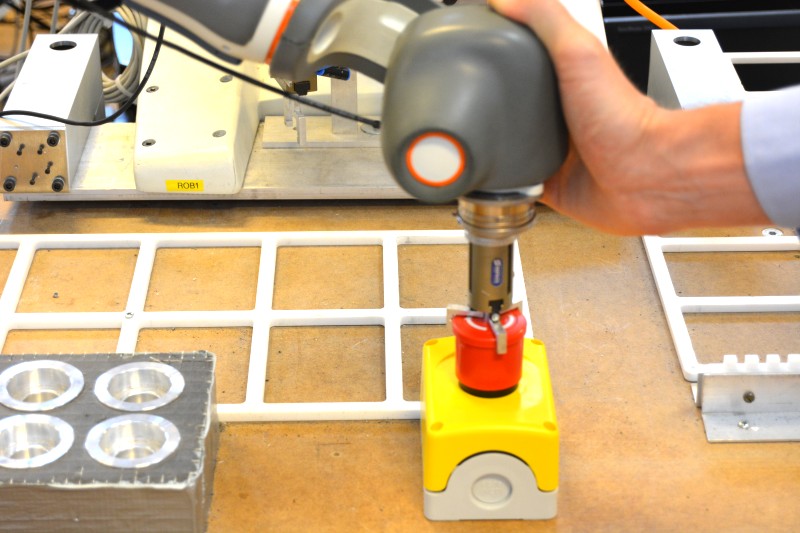}\subcaption{}
\vspace{0mm}
\end{minipage}
\hfill
\begin{minipage}{.48\columnwidth}
\centering
\includegraphics[width=\columnwidth]{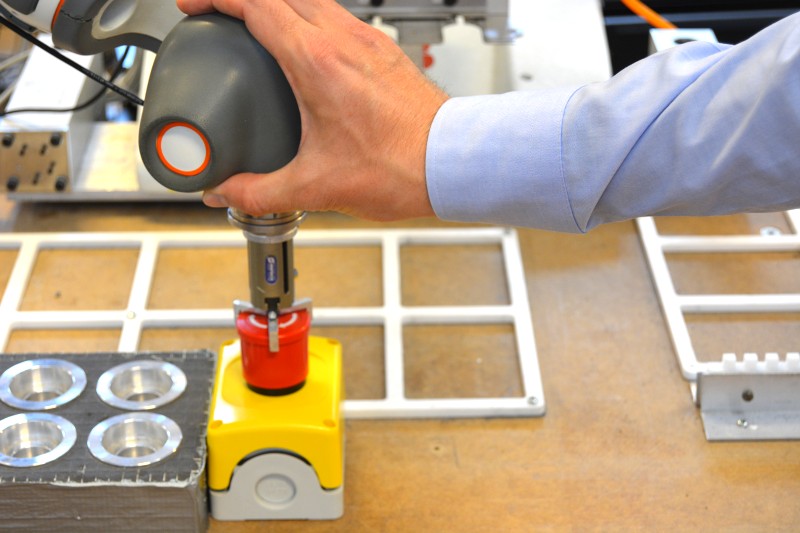}\subcaption{}
\vspace{0mm}
\end{minipage}
\caption{Photographs of Setup~1. The demonstration started as shown in (a), with the stop button grasped by the robot. The robot arm was moved by means of lead-through programming, first forward (b) from the robot's perspective, and subsequently downward until contact between the yellow case and the corresponding gray case, thus assembling the parts (c). Finally, the robot arm was moved sideways until contact between the assembled stop button and the environment (d).}
\label{fig:setup1}
\end{figure}

\begin{figure}
\begin{minipage}{.48\columnwidth}
\centering
\includegraphics[width=\columnwidth]{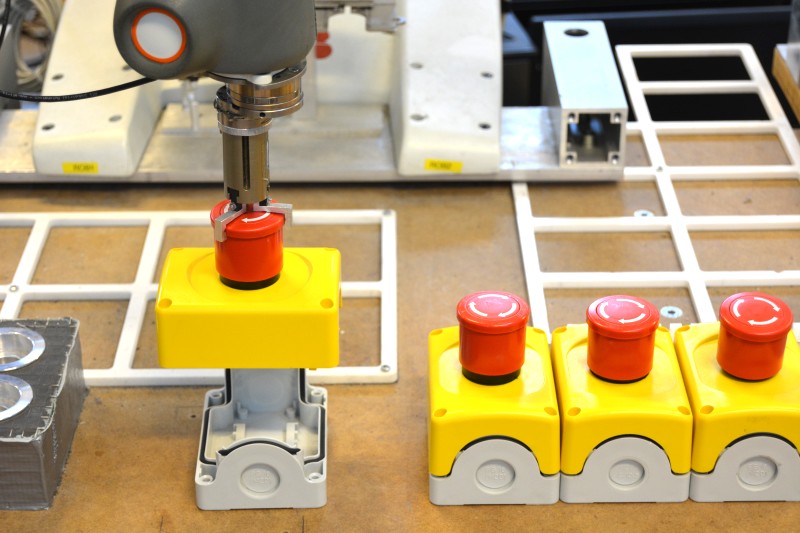}\subcaption{}
\vspace{3mm}
\end{minipage}
\hfill
\begin{minipage}{.48\columnwidth}
\centering
\includegraphics[width=\columnwidth]{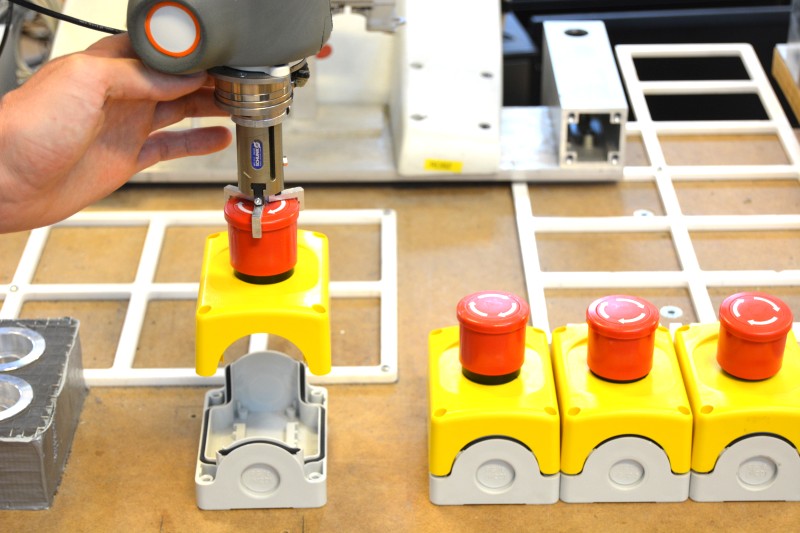}\subcaption{}
\vspace{3mm}
\end{minipage}
\begin{minipage}{.48\columnwidth}
\centering
\includegraphics[width=\columnwidth]{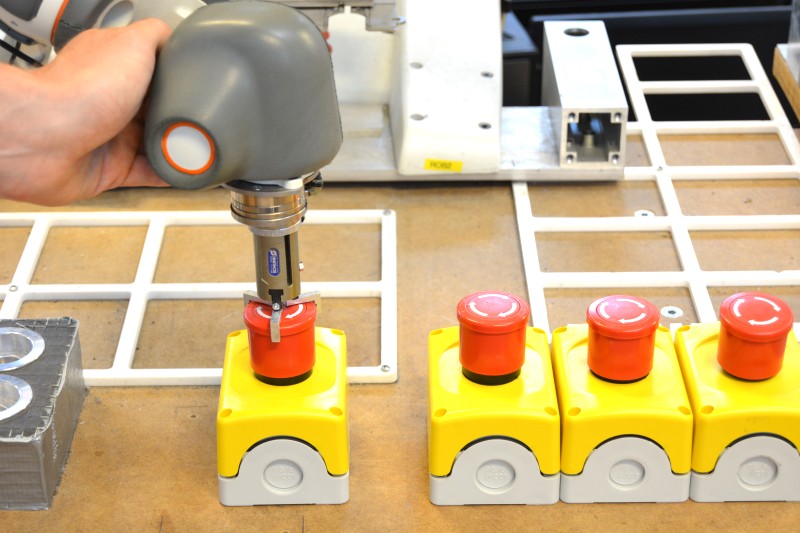}\subcaption{}
\vspace{3mm}
\end{minipage}
\hfill
\begin{minipage}{.48\columnwidth}
\centering
\includegraphics[width=\columnwidth]{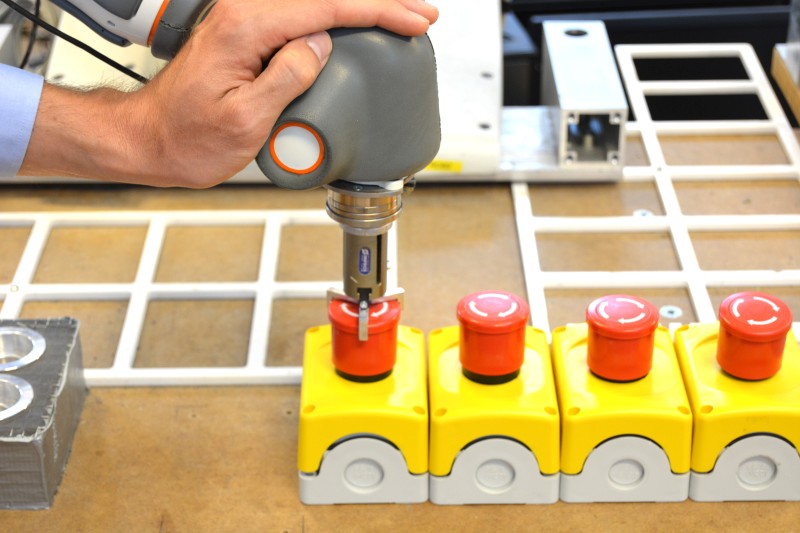}\subcaption{}
\vspace{3mm}
\end{minipage}
\begin{minipage}{.48\columnwidth}
\centering
\includegraphics[width=\columnwidth]{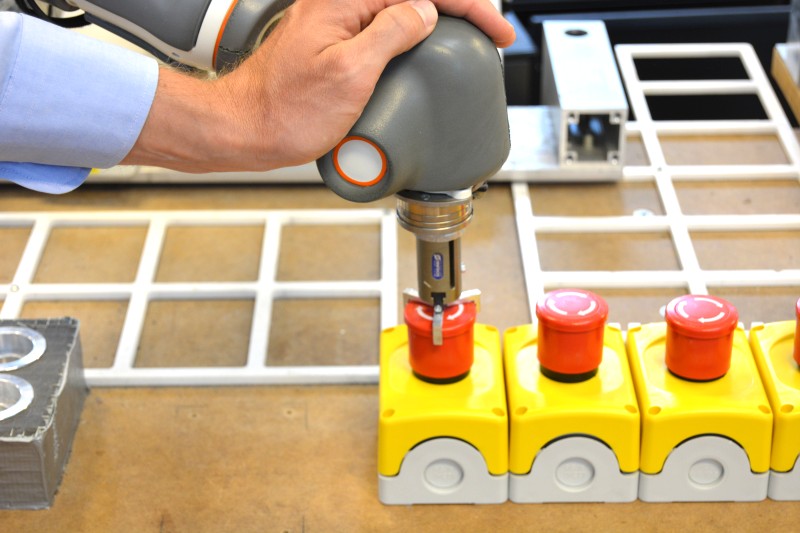}\subcaption{}
\vspace{0mm}
\end{minipage}
\hfill
\begin{minipage}{.48\columnwidth}
\centering
\includegraphics[width=\columnwidth]{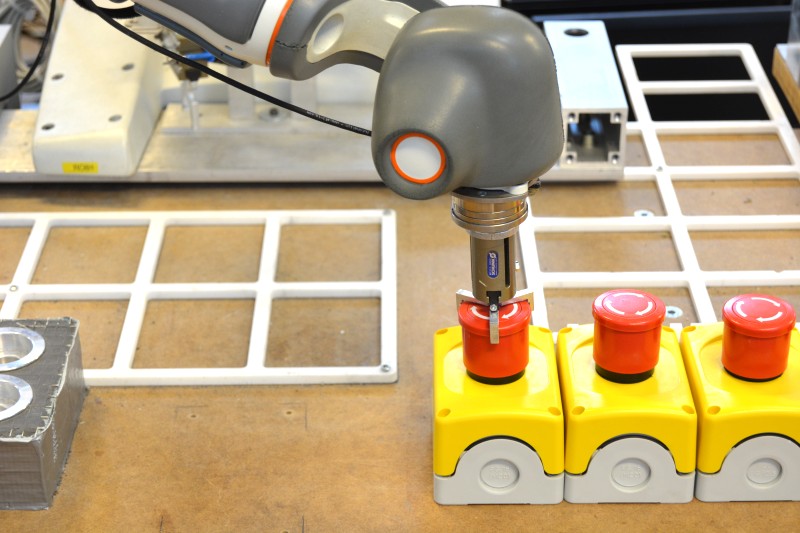}\subcaption{}
\vspace{0mm}
\end{minipage}
\caption{Photographs of Setup~2. The initial configuration is shown in (a). Previously assembled stop buttons are shown to the right. The robot gripper was first rotated to align the stop-button cases with each other (b), and then the grasped part was moved down until contact, so that the two cases were assembled (c). Thereafter, the robot arm was moved sideways along the table. In (d), a new contact was established between the grasped stop button and the ones that were previously assembled. The movement continued sideways (e) until the final configuration was reached (f).}
\label{fig:setup2}
\end{figure}

In this paper, we continue the research proposed in \cite{lee2015autonomous}, by extending the position-based approach to also take interaction forces into account. In robot tasks, it is common that the type of physical contact between the robot and its work objects changes during the transition between key phases. Examples of such transitions, in the context of robotic assembly, are shown in \cref{fig:setup1,fig:setup2}. In this paper, we use such changes in the measured interaction forces to detect these transitions. The approach in \cite{lee2015autonomous} was used to determine preliminary segmentation points based on demonstrated position data. A Kalman filter was used to detect sudden changes in the contact force measurements, and the time instances of these were used to verify and modify the initial set of segmentation points. In terms of contribution, the main component is described in \cref{sec:force_based}, and put into context in \cref{sec:method}.

Previous methods for segmentation based on force measurements were commonly based on hidden Markov models (HMMs) \cite{hannaford1991hidden,hundtofte2002building}. Our proposed method offers an alternative to HMM approaches, which is easy to implement and analyze, while retaining the benefits of the GMM-based segmentation in \cite{lee2015autonomous}. In addition to the segmentation itself, the force segmentation provides conditions for when to transit between segments during execution.

\section{Method}
\label{sec:method}
In this section, it is shown how to segment a demonstrated robot trajectory using both position and force data. This method extends the position-based autonomous segmentation framework in \cite{lee2015autonomous} by also including force data. The overall strategy is to apply the GMM segmentation described in \cite{lee2015autonomous}, and then use force data to verify segmentation points of the GMM, and possibly add new ones. Once a demonstrated trajectory has been segmented, each segment can be encoded as, for instance, a dynamical movement primitive (DMP), which in turn can be executed by robots. The novel segmentation component is described in \cref{sec:force_based}, whereas the remaining parts of this section describe how to combine it with existing frameworks.

\subsection{Position-Based Segmentation Using a GMM}
\label{sec:position_based}
For segmentation using position data, we apply the algorithm proposed in \cite{lee2015autonomous}, where a demonstrated trajectory is clustered into a GMM. In the following, the approach is briefly described for convenience, while a complete description can be found in \cite{lee2015autonomous}. A GMM consists of a set of Gaussian distributions, denoted $N(\xi|\mu_i, \Sigma_i)$, and we represent the mean vector $\mu_i$ and covariance matrix $\Sigma_i$ of each Gaussian as follows. 

\begin{equation}
\mu_i =
\begin{pmatrix}
\mu_{i,t} \\
\mu_{i,\xi}
\end{pmatrix}
\end{equation}
\begin{equation}
\Sigma_i =
\begin{pmatrix}
\Sigma_{i,t} & \Sigma_{i,t\xi} \\
\Sigma_{i,\xi t} & \Sigma_{i,\xi}
\end{pmatrix}
\end{equation}
Here, $t$ denotes the time and $\xi$ denotes the configuration of the robot. Further, $i$ is the index of the Gaussian, which we assume is ordered chronologically, so that the Gaussian with lowest temporal mean $\mu_{i,t}$ corresponds to the first index.

Each Gaussian of the GMM corresponds to a segment. An example is given in \cref{fig:gmm_trial1}, where a trajectory has been clustered into five segments. Each ellipse corresponds to the standard deviation of the corresponding Gaussian. 

In \cite{lee2015autonomous}, two alternative segmentation approaches are presented; one based on a geometrical interpretation of the Gaussians, and one using the GMM weights along the time component. For both approaches, segmentation time points are identified along the time component, where the standard deviations of any two adjacent Gaussians overlap, or where there is a space between them. This corresponds to one segmentation point between each of the ellipses in, \textit{e.g.}, \cref{fig:gmm_trial1}. These approaches give remarkably similar results \cite{lee2015autonomous}, and in this paper, we use the version based on the geometrical interpretation. Then, candidate time intervals for the segmentation points are identified between each two consecutive Gaussians, defined by the time boundaries
\begin{equation}
t_{\text{b}1} = \mu_{i,t} + \sqrt{\Sigma_{i,t}}
\end{equation}
and
\begin{equation}
t_{\text{b}2} = \mu_{(i+1),t} - \sqrt{\Sigma_{(i+1),t}}
\end{equation}
Subsequently, segmentation time points are established within these short time intervals; see \cite{lee2015autonomous}.

\subsection{Force-Based Segmentation Using Kalman Filtering}
\label{sec:force_based}
In this section, we use force data to assign segmentation points when there are significant changes in the contact between the robot and its environment. For instance, a transfer from free-space robot motion to contact with the surroundings, or \textit{vice versa}, results in a sudden change of the contact forces. Further, it is typically beneficial to distinguish between the different contact situations, by assigning them into different segments. In this paper, we use a Kalman filter (see Chapter 11 in \cite{aastrom2013computer}) to detect sudden changes in the force measurements, and define segmentation points when these changes occur.

Consider the following sampled state-space model for the contact force $F$.
\begin{align}
F(t+1) &= F(t) + v(t) \label{eq:ss1}\\
F_\text{m}(t) &=  F(t) + e(t) \label{eq:ss2}
\end{align}
Here, $t$ denotes a discrete time step, $F$ is the true contact force, $F_\text{m}(t)$ is the measurement of $F$ influenced by noise, and $v$ and $e$ are process noise and measurement noise, respectively. It is assumed that the noise components are discrete-time zero-mean Gaussian white-noise processes, uncorrelated with each other. Further, we introduce two noise covariance matrices as follows.
\begin{align}
R_1 &= E[v(t)v^\text{T}(t)]\\
R_2 &= E[e(t)e^\text{T}(t)]
\end{align}
Based on the state-space model, we define a Kalman filter as specified in \cite{aastrom2013computer}, where more information about general Kalman filtering  can be found. Because the system matrix and the mapping from state to measurements (disregarding the noise) in the state-space model are given by identity matrices, we obtain a Kalman filter that is simple as compared to the general case.
\begin{align}
&\hat{F}(t\;|\;t) = \hat{F}(t\;|\;t-1) + K_f(t)(F_\text{m}(t) - \hat{F}(t\;|\;t-1)) \\
&\hat{F}(t+1\;|\;t) = \hat{F}(t\;|\;t)
\end{align}
Here, $\hat{F}$ represents estimated force, and the conditional representation of, for example, $\hat{F}(t\;|\;t-1))$, specifies the available measurements, in this example measurements up to $F_\text{m}(t-1)$. The gain $K_f$ is given by
\begin{equation}
K_f(t) = P(t\;|\;t-1)(P(t\;|\;t-1) + R_2)^{-1}
\end{equation}
where $P$ is the variance of the estimation error and given by
\begin{gather}
\begin{aligned}
P(t+1\;|\;t) &= P(t\;|\;t-1) + R_1 \\
&\hspace{-2mm}-K_f(t)(P(t\;|\;t-1) + R_2)K_f^\text{T}(t)\\
P(t\;|\;t) &= P(t\;|\;t-1)\\
&\hspace{-2mm}-P(t\;|\;t-1)(P(t\;|\;t-1) + R_2)^{-1}P(t\;|\;t-1)\\
P(t=0) &= R_2
\end{aligned}
\end{gather}
The model given by \cref{eq:ss1,eq:ss2} is accurate only as long as the contact situation remains the same. However, significant contact changes are not well captured by the model, and these changes are therefore expected to yield a significant force estimation error. In the following, we omit the time variable  to keep the notation uncluttered, knowing that each entity is time dependent and assuming that measurements up to time $t$ (inclusive) are available. Let us define the error of the force estimate as
\begin{equation}
\tilde{F} = F - \hat{F}
\end{equation}
We use the estimation error, normalized with respect to its variance, as a measure of sudden changes in the contact forces. This measure is given by $\tilde{F}^\text{T}P^{-1}\tilde{F}$; see \cref{fig:kalman_setup1} for an example where this measure has been plotted over time. When $\tilde{F}^\text{T}P^{-1}\tilde{F}$ has a peak, this indicates a sudden change in the contact forces and therefore a transition between different kinds of contacts. Hence, segmentation points are assigned to times where $\tilde{F}^\text{T}P^{-1}\tilde{F}$ peaks.

\subsection{Combined Position and Force Segmentation}
The position-based GMM segmentation in \cref{sec:position_based} and \cite{lee2015autonomous} is enhanced by the force-based segmentation in \cref{sec:force_based} as follows. An initial set of segmentation points is determined from position data only; see \cref{sec:position_based}. Thereafter, each force-based segmentation point (see \cref{sec:force_based}) is used to adjust the initial set of segmentation points as follows:
\begin{itemize}
\item If an initial segmentation point is close to the force-based segmentation point, it is updated to the force-based segmentation point. This is motivated by the fact that the force data can provide segmentation points with high time resolution (see \cref{fig:kalman_setup1}) as compared to position data (see upper plot in \cref{fig:pos_force_setup1} and \cref{fig:gmm_trial1}).
\item If the force-based segmentation point is not close to any initial segmentation point, it is added as a new segmentation point. Such a situation may occur when the contact situation is changed, without affecting the robot movement significantly. 
\end{itemize}
These updates transform the initial set of segmentation points, into a final set.

\subsection{Representation and Autonomous Execution of Segmented Demonstration}
We use DMPs \cite{ijspeert2013dynamical} with temporal coupling to represent and execute the movements in operation space. DMPs are motion control laws, developed to enhance robot's replanning capabilities, and commonly used in the context of programming by demonstration \cite{niekum2015learning,karlsson2017autonomous,karlsson2017motion,chiara2018passivity}. For configurations in real coordinate systems $\mathbb{R}^n$, DMPs are commonly based on the following dynamical system.
\begin{equation}
\label{eq:dmp_main1}
\tau^2 \ddot{y}_\text{r} = \alpha(\beta(y_g-y)-\tau \dot{y}) + f(x)
\end{equation}
Here, $\ddot{y}_\text{r}$ is a reference acceleration sent to the internal robot controller, $y$ denotes robot configuration, $y_g$ is the goal configuration, $\alpha$ and $\beta$ are positive constants, $\tau$ is a positive time parameter, and $f(x)$ is a so-called forcing term where $x$ is a phase variable \cite{ijspeert2013dynamical}. The DMP formulation in \cref{eq:dmp_main1} can be extended to also incorporate control of orientation in operation space, as explained in \cite{ude2014orientation}. Further, temporal coupling \cite{ijspeert2013dynamical,karlsson2017dmp,karlsson2018convergence} was included to support online replanning.

Given a segmented demonstrated trajectory, we use the position data in each segment to determine a corresponding DMP. If the end point of a given segment corresponds to a contact change, an offset is added to move the goal configuration in the movement direction, and a force threshold is used to detect movement completion during execution. If the end point does not correspond to a contact change, the movement is considered completed when $y$ is close to $y_g$. In summary, a demonstration is first divided into segments, and thereafter each segment is represented as a DMP. To replay a demonstration autonomously, the corresponding sequence of DMPs is executed.

\section{Experiments}
To validate the proposed method, a prototype of the ABB YuMi robot was used; see \cref{fig:yumi_main} and \cite{yumi}. Each wrist of the robot was equipped with an ATI \cite{ati_mini} Mini40 force/torque sensor. The sensorless lead-through programming implemented in \cite{stolt2015sensorless} was used for demonstrating trajectories. During the lead-through programming, the robot was grasped so that the contact with the operator was not directly measured by the force/torque sensor. The method was implemented on an ordinary PC, which communicated with the internal robot controller through the research interface ExtCtrl \cite{blomdell2005extending,blomdell2010flexible}, running at \SI{250}{Hz}. Two different setups were considered to investigate the segmentation.

\textbf{Setup~1} The task of the robot was to assemble two parts of an emergency stop button; see \cref{fig:setup1}. This was done by first moving the grasped part in the $x$-direction, and subsequently moving downwards until contact was established, which accomplished the assembly. Subsequently, the assembled parts were moved sideways in a sliding motion, until a second contact was established.

\textbf{Setup~2} Again, the task was to assemble two parts of an emergency stop button. This setup is visualized in \cref{fig:setup2}. First, the grasped part was rotated to align it with the part on the table. Thereafter, it was moved downward until contact, in the same way as for Setup~1. The assembled parts were then moved sideways. During the movement, contact was established with previously assembled stop buttons, so that these were also moved to the side.

In Setup~1, we considered only position and force data, which is sufficient to demonstrate the key features of the segmentation process. Setup~2 also included orientation and torque data, to show how these data can be incorporated in the segmentation. After demonstration and segmentation, each task was executed autonomously on the robot.

A video that shows the experiments is available as an attachment to this paper, and online \cite{segmentation_youtube}.

\begin{figure}
	\centering	
	\includegraphics[width=8cm]{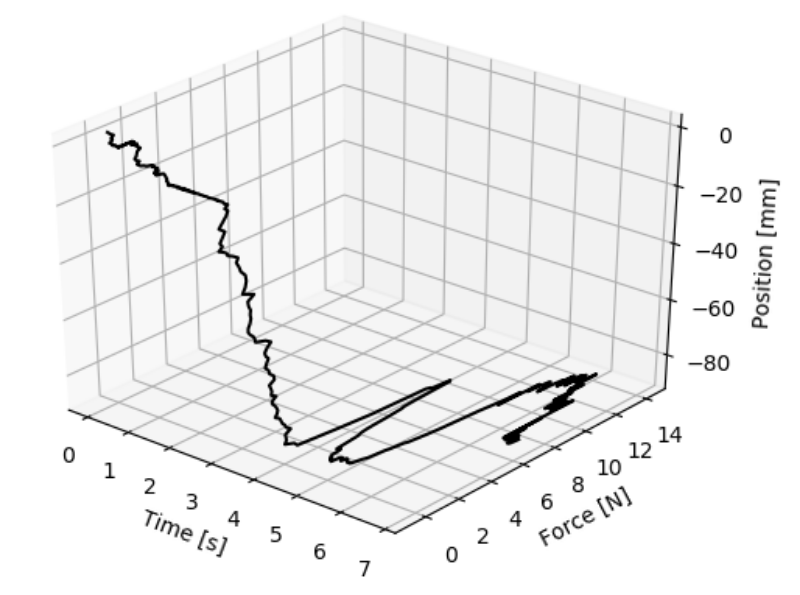}
	\caption{Data from Setup~1. This graph shows position and force from the demonstration, along the vertical axis ($z$). The downward movement began at $t\approx$ \SI{3}{s}. As expected, there were no contact forces during this free-space movement. When contact was established, the downward movement stopped, and it can be seen that the normal force increased at that point.}
	\label{fig:plot_3d}
\end{figure}

\begin{figure}
	\centering
	\input{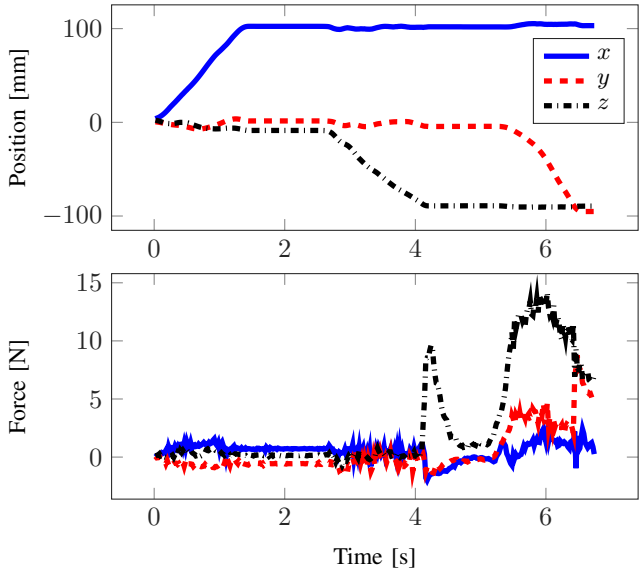}
	\caption{Measurements from Setup~1. The upper plot shows position, and one movement in each of the $x$, $z$, and $y$-directions are clearly visible. The lower plot shows contact forces. A sudden change can be distinguished at $t \approx$ \SI{4}{s}, but apart from that, it is not obvious how to assign segmentation points only by inspecting the raw force data.}
	\label{fig:pos_force_setup1}
\end{figure}

\begin{figure}
	\centering	
	\includegraphics[width=9cm]{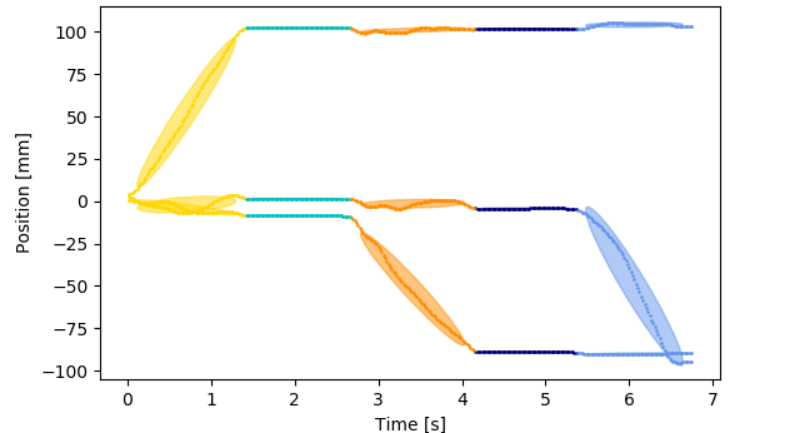}
	\caption{Position data from Setup~1, also shown in the upper plot in \cref{fig:pos_force_setup1}, clustered as a GMM with five Gaussians. Each color corresponds to one segment. The movements in the $x$, $z$, and $y$-directions have been divided into one cluster each. In addition, the two parts of the trajectory with zero velocity have been divided into one cluster each.}
	\label{fig:gmm_trial1}
\end{figure}

\begin{figure}
	\centering	
	\input{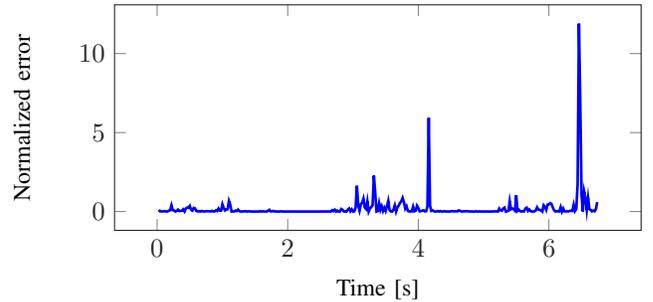}
	\caption{Normalized force estimation error of the Kalman filter, from Setup~1. Two peaks are clearly visible. These peaks correspond to new contact between the robot and its surroundings. This can be compared with the lower plot in \cref{fig:pos_force_setup1}, where these changes are not as easy to distinguish.}
	\label{fig:kalman_setup1}
\end{figure}

\section{Results}
\label{sec:results}
Data from Setup~1 are shown in \cref{fig:plot_3d,fig:pos_force_setup1,fig:gmm_trial1,fig:kalman_setup1}. The demonstrated trajectory and contact forces are shown in \cref{fig:pos_force_setup1}. These data were used as input for the segmentation algorithm. The initial segmentation, obtained as described in \cref{sec:position_based}, is shown in \cref{fig:gmm_trial1}. The normalized force estimation error, obtained according to \cref{sec:force_based}, is shown in \cref{fig:kalman_setup1}. Two peaks are clearly visible, and these correspond to the establishment of new contacts; first between the parts to be assembled and subsequently with another object in the work space. The instances of these peaks were used to verify and adjust the segmentation points at $t\approx$ \SI{4}{s} and $t\approx$ \SI{6.5}{s}, initially obtained according to \cref{fig:gmm_trial1}. The segmentation resulted in five segments, each of which has a clear semantic interpretation:
\begin{enumerate}
\item Move end effector forward (from the perspective of the robot);
\item Idle (zero velocity);
\item Move down until contact;
\item Idle;
\item Move sideways until contact.
\end{enumerate}
Idle segments can emerge naturally from lead-through demonstrations, for instance if the operator stops to change grasp. For many tasks, these idle segments can be omitted during execution; see \cref{sec:discussion}.
This was also done for the execution phases in these experiments.

\begin{figure}
	\centering
	\input{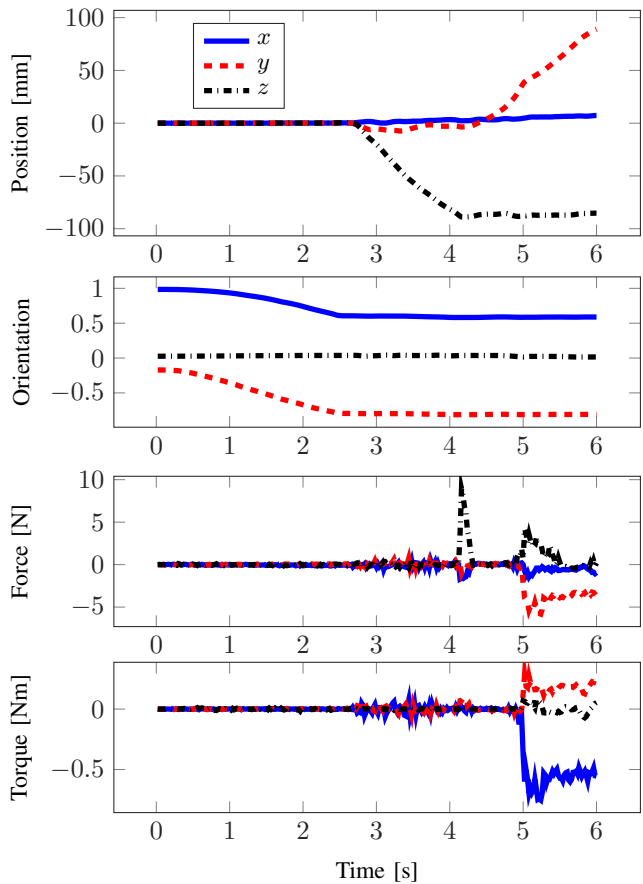}
	\caption{Measurements from Setup~2. The rotation (see the second plot from above) and the two translations (see the first plot from above) are clearly visible. Here, the orientation is represented by the imaginary part of the corresponding quaternion. The directions of the position, force, and torque are defined by the legend.}
	\label{fig:pos_force_setup2}
\end{figure}

\begin{figure}
\centering
\begin{minipage}{.96\columnwidth}
\centering
\includegraphics[width=9cm]{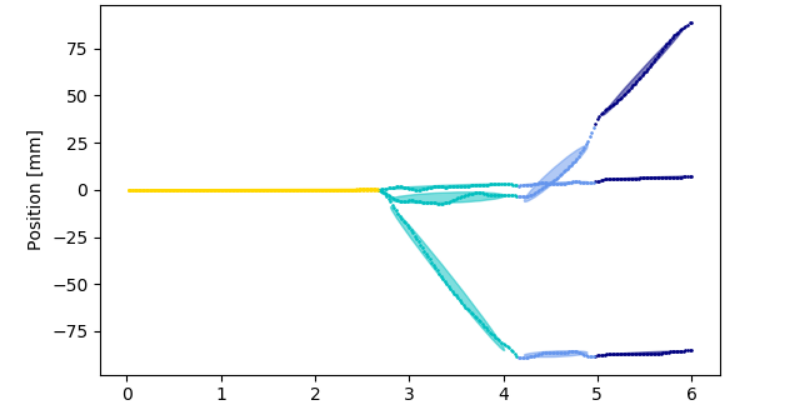}
\end{minipage}\hfill
\par\vspace{0mm} 
\begin{minipage}{.96\columnwidth}
\includegraphics[width=9cm]{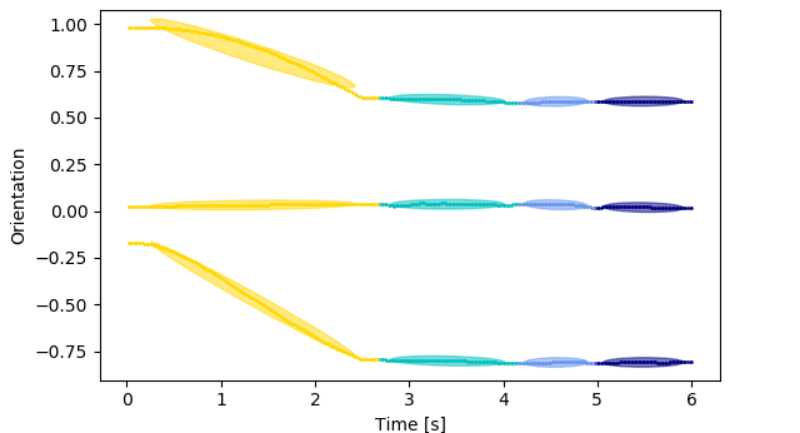}
\end{minipage}\hfill
\caption{Configuration data from Setup~2, also shown in the two upper plots in \cref{fig:pos_force_setup2}, clustered as a GMM with four Gaussians.}
\label{fig:gmm_setup2}
\end{figure}

\begin{figure}
	\centering	
\begin{tikzpicture}

\begin{axis}[
ylabel={Normalized error},
xlabel={Time [s]},
width=7cm,
height=3cm,
scale only axis,
every outer x axis line/.append style={white!15!black},
every x tick label/.append style={font=\color{white!15!black}},
every outer y axis line/.append style={white!15!black},
every y tick label/.append style={font=\color{white!15!black}},
]
\addplot [solid,line width=1.0pt, color=blue]
table {%
0.02 0
0.04 0.0028274897895661
0.06 0.00069511987182307
0.08 0.00103179194663961
0.1 0.00151705326662483
0.12 0.00555346433485334
0.14 0.00198778715790378
0.16 0.00360437678926358
0.18 0.0224384859450224
0.2 0.00616044360236691
0.22 0.0126286095288679
0.24 0.00420643190356504
0.26 0.0101327444496739
0.28 0.00358288205683208
0.3 0.0511020256408168
0.32 0.00109013502071583
0.34 0.0181346006018849
0.36 0.00244118389884137
0.38 0.0267603042819846
0.4 0.0379006702752657
0.42 0.0216871689384044
0.44 0.00460868695915201
0.46 0.00842958457908225
0.48 0.00420833023167568
0.5 0.0229618382436757
0.52 0.00587757557185783
0.54 0.000953606407866423
0.56 0.0151595139568938
0.58 0.0147445292287085
0.6 0.0282822221680375
0.62 0.0189915690865757
0.64 0.00839495684794891
0.66 0.0168991282324423
0.68 0.00969710667491825
0.7 0.00541904695795133
0.72 0.0116837714405839
0.74 0.00125109661733851
0.76 0.0024811615847964
0.78 0.0125731132560581
0.8 0.00451188238603631
0.82 0.00901947353408895
0.84 0.0116903056228247
0.86 0.00401630349089736
0.88 0.00435201022587153
0.9 0.023488531963541
0.92 0.0103818737206664
0.940000000000001 0.0175601680088175
0.960000000000001 0.00329460614778853
0.980000000000001 0.00620599708176186
1 0.00702835157212081
1.02 0.0110714255822131
1.04 0.00515411365524781
1.06 0.00122210222016007
1.08 0.00206210569488601
1.1 0.00431159470008474
1.12 0.0052487656401179
1.14 0.0147817869633832
1.16 0.0190284110274077
1.18 0.0104949942401187
1.2 0.0104326653578449
1.22 0.00145451542806415
1.24 0.0049981022823448
1.26 0.00220422405217612
1.28 0.00818792202905448
1.3 0.00224962546460024
1.32 0.00487404612289578
1.34 0.00490832649150916
1.36 0.00598377617245315
1.38 0.00665033085725076
1.4 0.0114286745802172
1.42 0.00516898069158024
1.44 0.0164566489934841
1.46 0.000267101936240104
1.48 0.0245646471747391
1.5 0.0124987283982758
1.52 0.0209180341773553
1.54 0.0167008394019702
1.56 0.00687361895011161
1.58 0.0131672888988686
1.6 0.0541331475155116
1.62 0.0323852686675116
1.64 0.0539450704174173
1.66 0.0221467923683969
1.68 0.00838285520565008
1.7 0.0222770639971968
1.72 0.00338266246716327
1.74 0.0622076825162283
1.76 0.0173984785372238
1.78 0.00439763665323236
1.8 0.00205864142336673
1.82 0.0175220070350609
1.84 0.0128433813389362
1.86 0.0270978632865792
1.88 0.0144545302329441
1.9 0.024019964572153
1.92 0.0237584063855959
1.94 0.0191721029550045
1.96 0.0512856173255993
1.98 0.0313933617318794
2 0.00987474293997213
2.02 0.0104795416937525
2.04 0.052947665155388
2.06 0.0212281984123835
2.08 0.016648384134117
2.1 0.0066102161784303
2.12 0.027842218856543
2.14 0.00592979877450335
2.16 0.0123050341654203
2.18 0.00465379411020528
2.2 0.0288645695378368
2.22 0.00669621345018384
2.24 0.0420848652326468
2.26 0.0132182044623412
2.28 0.0158258687546956
2.3 0.0161212101533055
2.32 0.018437953141035
2.34 0.004895241308878
2.36 0.0293658323129607
2.38 0.0298999584573498
2.4 0.0209112267955661
2.42 0.0288178169080139
2.44 0.0280306277741377
2.46 0.0121204023983493
2.48 0.0063677006562008
2.5 0.00320729535613884
2.52 0.0122648208945416
2.54 0.00170060902985813
2.56 0.00217719853353422
2.58 0.000560014115401264
2.6 0.0109748253848153
2.62 0.0347119328791359
2.64 0.0464257651817188
2.66 0.0302162789621569
2.68 0.175876342835147
2.7 0.0578615675255152
2.72 0.0165786141977732
2.74 0.0905532472081048
2.76 0.0647084365222431
2.78 0.088531346501083
2.8 0.23789506618614
2.82 0.0279721841919693
2.84 0.5300991137934
2.86 0.268647871851162
2.88 0.493119500666518
2.9 0.0505488458943546
2.92 0.124109345242674
2.94 0.0373356958823526
2.96 0.294247280233927
2.98 0.241790582795283
3 0.369316906085871
3.02 0.402728862071593
3.04 0.293797838828085
3.06 0.132877152699584
3.08 0.13924059773104
3.1 0.787811588652729
3.12 0.696498840430528
3.14 0.141389417867167
3.16 0.136996595591583
3.18 0.0135543583807162
3.2 0.0344934726221746
3.22 0.135940346496289
3.24 0.172335768796884
3.26 0.391993239388344
3.28 0.154778757795087
3.3 0.140658737721489
3.32 0.376288760353927
3.34 0.0958829362503422
3.36 0.00187204921042945
3.38 0.29309525586066
3.4 0.0732396940719651
3.42 0.516223473914518
3.44 0.710236771001479
3.46 1.78806310640739
3.48 0.577437420197579
3.5 1.29708101564632
3.52 0.685153345491093
3.54 1.1925174301072
3.56 0.339707715306931
3.58 0.560472845603879
3.6 0.219245214785049
3.62 0.181070993808898
3.64 0.13337631837046
3.66 0.539994537195495
3.68 0.30455671994604
3.7 0.14374943236941
3.72 0.829089221597211
3.74 1.31850920631333
3.76 0.906535431823331
3.78 0.138698880859712
3.8 0.201282031365138
3.82 0.51470242384026
3.84 0.458209543053052
3.86 0.241826980356301
3.88 0.165795406936967
3.9 0.034278071654248
3.92 0.287726080728182
3.94 0.0768428176104586
3.96 0.0106285576660744
3.98 0.0756923406285436
4 0.0395254783100577
4.02 0.0223540889183432
4.04 0.0247683808317795
4.06 0.131247020692869
4.08 0.264708340264622
4.1 0.295776742198956
4.12 0.181975111958923
4.14 20.4223358533224
4.16 46.2873964798958
4.18 0.0785355167026372
4.2 2.34768903707167
4.22 0.945812055573246
4.24 7.49798427459033
4.26 1.42044911286437
4.28 4.40774761079137
4.3 2.2612485445425
4.32 1.45442627490226
4.34 0.0609656286099567
4.35999999999999 0.0150493950077873
4.37999999999999 0.0141737717497493
4.39999999999999 0.00728182602579985
4.41999999999999 0.0111785026194861
4.43999999999999 0.0160029065555947
4.45999999999999 0.00759539843177837
4.47999999999999 0.0208971784306326
4.49999999999999 0.0178703192347156
4.51999999999999 0.0412239367619097
4.53999999999999 0.0211565711494163
4.55999999999999 0.0137118878609118
4.57999999999999 0.105769837509229
4.59999999999999 0.0364332489151301
4.61999999999999 0.0213311777786022
4.63999999999999 0.0206319451542847
4.65999999999999 0.0314490994279353
4.67999999999999 0.00779104709853749
4.69999999999999 0.0183336566504369
4.71999999999999 0.0199967007266626
4.73999999999999 0.0271913137822297
4.75999999999999 0.0292350679467057
4.77999999999999 0.0243868926138646
4.79999999999999 0.167562547301276
4.81999999999999 0.0940780245859045
4.83999999999998 0.102446273030158
4.85999999999998 0.172056436887609
4.87999999999998 0.563050423990925
4.89999999999998 0.165729593865018
4.91999999999998 2.06574582475417
4.93999999999998 0.81359357971791
4.95999999999998 0.175190673461026
4.97999999999998 0.613163498325964
4.99999999999998 6.58809605708688
5.01999999999998 20.0603108003555
5.03999999999998 0.882779529385134
5.05999999999998 1.69609997324854
5.07999999999998 2.46297415527509
5.09999999999998 0.635894987422367
5.11999999999998 0.165596391770714
5.13999999999998 1.25921127071126
5.15999999999998 0.681230794310804
5.17999999999998 1.82959359458835
5.19999999999998 0.915640572629101
5.21999999999998 0.123126789773867
5.23999999999998 0.0406648785393106
5.25999999999998 1.3126512630718
5.27999999999998 4.01254669454134
5.29999999999997 0.112815816755427
5.31999999999997 0.357502334475717
5.33999999999997 0.147596000341065
5.35999999999997 2.27393953846719
5.37999999999997 0.44104377760753
5.39999999999997 0.822325000391776
5.41999999999997 0.357126332788755
5.43999999999997 1.17029902206532
5.45999999999997 0.0342244517957635
5.47999999999997 0.934637701838519
5.49999999999997 0.30091183210304
5.51999999999997 0.34769581794794
5.53999999999997 0.0106207501677698
5.55999999999997 0.086979778650167
5.57999999999997 0.11285436947876
5.59999999999997 0.0263264607097646
5.61999999999997 0.0917759567874007
5.63999999999997 0.0434855295571947
5.65999999999997 0.0226660715660625
5.67999999999997 0.1490898269239
5.69999999999997 0.0281795772971924
5.71999999999997 0.00308285445239408
5.73999999999997 0.0430356026954431
5.75999999999997 0.101514075098754
5.77999999999996 0.0737770087099388
5.79999999999996 0.128789698667539
5.81999999999996 0.0519501675595389
5.83999999999996 0.0297356954387191
5.85999999999996 0.148815322863608
5.87999999999996 0.12895986284656
5.89999999999996 0.276863549184622
5.91999999999996 0.36646702010835
5.93999999999996 0.284120284519734
5.95999999999996 0.349680050039471
5.97999999999996 0.457979776185816
5.99999999999996 0.178202746087737
};
\end{axis}

\end{tikzpicture}
	\caption{Normalized force estimation error of the Kalman filter, from Setup~2.}
	\label{fig:kalman_setup2}
\end{figure}
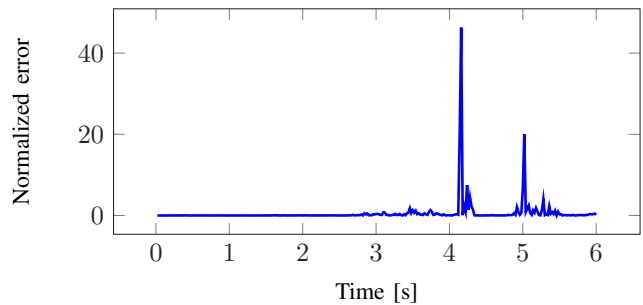

Data from Setup~2 are shown in \cref{fig:pos_force_setup2,fig:gmm_setup2,fig:kalman_setup2}. Similarly to Setup~1, the two peaks in \cref{fig:kalman_setup2} were used to verify and adjust two of the initial segmentation points in \cref{fig:gmm_setup2}; at $t\approx$ \SI{4}{s} and $t\approx$ \SI{5}{s}. These two segmentation points correspond to established contact between the assembly parts, followed by contact between the grasped stop button and other previously assembled stop buttons. Again, the method provided segments that can be interpreted semantically:
\begin{enumerate}
\item Rotate end effector;
\item Move down until contact;
\item Move sideways until contact, and continue movement;
\item Continue movement, retaining contact with table and previously assembled stop buttons.
\end{enumerate}

The autonomous execution of each demonstrated task was successful, and is shown in the video \cite{segmentation_youtube}.

\section{Discussion}
\label{sec:discussion}
In the experiments, the proposed method successfully segmented the demonstrated tasks into intuitively meaningful sub-tasks. It also provided pre- and post conditions for the segments, \textit{i.e.}, transit at a demonstrated configuration, or transit upon new contact. The segmented demonstrations allowed for the robot to perform the tasks autonomously.

Omitting idle examples, which was discussed in \cref{sec:results}, is an example of rearrangement of segments for task modification. More such examples are available in \cite{lee2015autonomous}, where a humanoid was used for a rice-cooking task and a food-cutting task. Segments were repeated and rearranged to perform different sub-tasks variable amounts of times. The ability of such updating and debugging, which would be very difficult to achieve using raw data only, is one of the main benefits of successful segmentation.

One obstruction when developing and evaluating segmentation methods, is that it is partly subjective whether a given segmentation result is successful or not. It is dependent on the user and the context. In this paper, a clear semantic meaning of the segments is considered a benefit. This criteria is subjective. Further, the segmentation should enable successful autonomous execution of the demonstrated task. This criteria can be tested objectively. An interesting direction of future work would be to use the segments as prior information in RL. It would then be interesting to investigate, if more objective criteria could be specified given that the RL should succeed within as few attempts as possible. Such criteria could be used both in the design and evaluation of future segmentation methods.

\section{Conclusion}
A method for autonomous segmentation of demonstrated robot tasks was proposed. The method in \cite{lee2015autonomous} was extended to also incorporate contact forces, which is the main contribution of this paper. Position trajectories were clustered into segments using GMMs \cite{lee2015autonomous}, and Kalman filtering was used to detect sudden changes in the contact forces. The method generated segments with clear semantic interpretations and transition criteria. The method was verified experimentally using an industrial robot. After demonstration, the segmented tasks could be performed autonomously by the robot. A video that shows the functionality is available as an attachment to this paper, and online \cite{segmentation_youtube}.

\bibliographystyle{IEEEtran}
\bibliography{IEEEabrv,gmm_kalman_segmentation}
\end{document}